\documentclass{amsart}

\usepackage{graphicx}

\title{Ontological Crises in Artificial Agents' Value Systems}
\author{Peter de Blanc \\ The Singularity Institute for Artificial Intelligence}
\date{\today}

\begin{document}
\begin{titlepage}
\maketitle
\end{titlepage}

\section{Abstract}
Decision-theoretic agents predict and evaluate the results of their actions using a model, or ontology, of their environment. An agent's goal, or utility function, may also be specified in terms of the states of, or entities within, its ontology. If the agent may upgrade or replace its ontology, it faces a crisis: the agent's original goal may not be well-defined with respect to its new ontology. This crisis must be resolved before the agent can make plans towards achieving its goals.

We discuss in this paper which sorts of agents will undergo ontological crises and why we may want to create such agents. We present some concrete examples, and argue that a well-defined procedure for resolving ontological crises is needed. We point to some possible approaches to solving this problem, and evaluate these methods on our examples.

\section{Introduction: Goals and Utility Functions}
An agent is any person or thing that performs actions in order to achieve a goal. These goals may involve anything of which the agent is aware, from its own inputs and outputs to distant physical objects. When creating an artificial agent, it is natural to be interested in which goals we choose to give it. When we create something, we usually do so because we expect it to be useful to us. Thus the goals we give to artificial agents should be things that we want to see accomplished.

Programmers of artificial agents are then faced with the task of specifying a goal. In our discussion we assume that goals take the form of utility functions defined on the set of possible states within the agent's ontology. If a programmer is specifying a utility function "by hand" -- that is, by looking at the ontology and directly assigning utilities to different states -- then the ontology must be comprehensible to the programmer. This will typically be the case for an ontology that the programmer has designed, but not necessarily so for one that an agent has learned from experience.

An agent with a fixed ontology is not a very powerful agent, so we would like to discuss an agent that begins with an ontology that its programmers understand and have specified a utility function over, and then upgrades or replaces its ontology. If the agent's utility function is defined in terms of states of, or objects within, its initial ontology, then it cannot evaluate utilities within its new ontology unless it translates its utility function somehow.

Consider, for example, an agent schooled in classical physics. Perhaps this agent has a goal that is easy to specify in terms of the movement of atoms, such as to maintain a particular temperature within a given region of space. If we replace our agent's ontology with a quantum one, it is no longer obvious how the agent should evaluate the desirability of a given state. If its utility function is determined by temperature, and temperature is determined by the movement of atoms, then the agent's utility function is determined by the movement of atoms. Yet in a quantum worldview, atoms are not clearly-defined objects. Atoms are not even fundamental to a quantum worldview, so the agent's ontology may contain no reference to atoms whatsoever. How then, can the agent define its utility function?

One way to sidestep the problem of ontological crises is to define the agent's utility function entirely in terms of its percepts, as the set of possible percept-sequences is one aspect of the agent's ontology that does not change. Marcus Hutter's universal agent AIXI \cite{AIXI} uses this approach, and always tries to maximize the values in its reward channel. Humans and other animals partially rely on a similar sort of reinforcement learning, but not entirely so.

We find the reinforcement learning approach unsatisfactory. As builders of artificial agents, we care about the changes to the environment that the agent will effect; any reward signal that the agent processes is only a proxy for these external changes. We would like to encode this information directly into the agent's utility function, rather than in an external system that the agent may seek to manipulate.

\section{Our Approach}

We will approach this problem from the perspective of concrete, comprehensible ontologies. An AI programmer may specify an ontology by hand, and then specify a utility function for that ontology. We will then try to devise a systematic way to translate this utility function to different ontologies.

When using this method in practice, we might expect the agent to have a probability distribution over many ontologies, perhaps specified concisely by the programmer as members of a parametric family. The programmer would specify a utility function on some concrete ontology which would be automatically translated to all other ontologies before the agent is turned on. In this way the agent has a complete utility function.

However, for the purposes of this discussion, we may imagine that the agent has only two ontologies, one old and one new, which we may call $O_0$ and $O_1$. The agent's utility function is defined in terms of states of $O_0$, but it now believes $O_1$ to be a more accurate model of its environment. The agent now faces an ontological crisis -- the problem of translating its utility function to the new ontology $O_1$.

In this paper we will present a method for addressing these problems. Our intention, however, is not to close the book on ontological crises, but rather to open it. Our method is of an ad-hoc character and only defined for a certain class of ontologies. Furthermore it is not computationally tractable for large ontologies. We hope that this discussion will inspire other thinkers to consider the problem of ontological crises and develop new solutions. 

\section{Finite State Models}

We wil nowl consider a specific kind of ontology, which we may call a \emph{finite state model}. These models have some finite set of possible hidden states, which the agent does not directly observe. On each time step, the model inputs some symbol (the agent's output), enters some hidden state, and outputs some symbol (the agent's input). The model's output depends (stochastically) only on its current state, while its state depends (stochastically) on both the input and the previous state.

\begin{center}
	\includegraphics[scale=0.5]{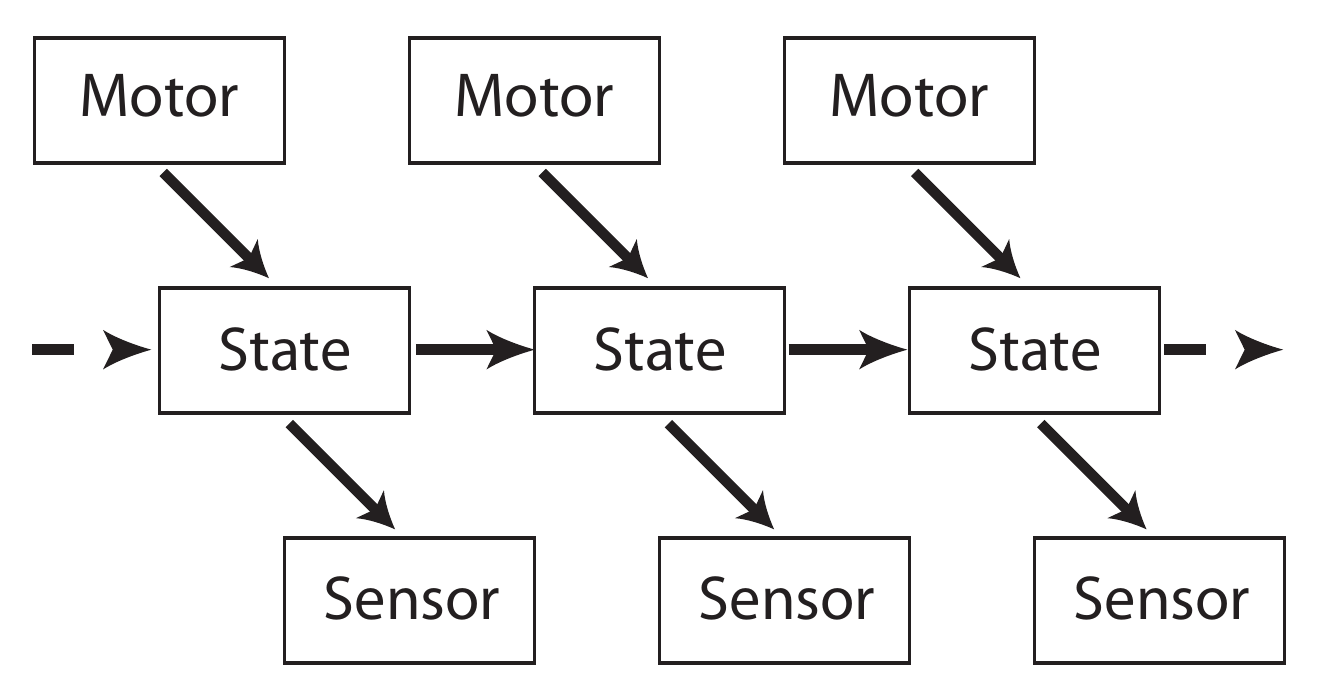}
\end{center}

Let us call the agent's output symbols \emph{motor symbols} and the agent's input symbols \emph{sensor symbols}. We will call the sets of symbols the motor alphabet and the sensor alphabet, denoted $M$ and $S$ respectively. We will assume that the alphabets are fixed properties of our agent's embodiment; we will not consider models with different alphabets.

Let $m = |M|$ and $s = |S|$. Then a model with $n$ states may be completely specified by $m$ different $n \times n$ transition matrices and one $s \times n$ output matrix.

For each $x \in M$, let us call the state transition matrix $T^x$. Note that the superscript here is not an exponent. We may call the output probability matrix $A$. Since we will be speaking of two ontologies, $O_0$ and $O_1$, we will use subscripts to indicate which ontology we are taking these matrices from; for instance, $T_0^x$ is the state transition matrix for action $x$ in the $O_0$ ontology.

\section{Maps between Ontologies}

Our basic approach to translating our utility function from $O_0$ to $O_1$ will be to construct a function from $O_1$ to $O_0$ and compose our utility function with this function. If

\begin{equation}U: O_0 \rightarrow \mathbb{R}\end{equation}

is a utility function defined on $O_0$, and $\phi: O_1 \rightarrow O_0$, then $U \circ \phi$ is a utility function defined on $O_1$.

The function $\phi$  we will seek to define will be a \emph{stochastic function}; its output will not be a single state within $O_0$, but a probability distribution over states. Thus if $O_0$ has $n_0$ states while $O_1$ has $n_1$ states, $\phi$ will be most naturally expressed as an $n_0 \times n_1$ matrix.

Let us consider some desiderata for $\phi$:

1. $\phi$ should be determined by the \emph{structure} of the models $O_0$ and $O_1$; the way in which the states are labeled is irrelevant.

2. If $O_0$ and $O_1$ are isomorphic to each other, then $\phi$ should be an isomorphism.

This may seem irrelevant, for if $O_1$ is isomorphic to $O_0$, then there is no need to change models at all. Nevertheless, few would object to 2 on grounds other than irrelevance, and 2 may be seen as a special case of a more general statement:

3. If $O_0$ and $O_1$ are \emph{nearly} isomorphic to each other, then $\phi$ should nearly be an isomorphism.

This criterion is certainly relevant; since $O_0$ and $O_1$ are both models of the same reality, they can be expected to be similar to that reality, and thus similar to each other.

In accordance with these desiderata, we will try to construct a function that is as much like an isomorphism as possible. To accomplish this, we will define in quantitative terms what we mean by "like an isomorphism." First, we observe that isomorphisms are invertible functions; thus, we will define a second function, which we fancifully call $\phi^{-1} : O_0 \rightarrow O_1$, even though it may not be a true inverse of $\phi$, and we will optimize both $\phi$ and $\phi^{-1}$ to be "like isomorphisms".

Our criterion is a combination of the computer science notion of \emph{bisimulation} with the information-theoretic idea of \emph{Kullback-Leibler divergence}.

Bisimulation means that either model may be used to simulate the other, using $\phi$ and $\phi^{-1}$ to translate states between models. Thus, for any action $x$, we would like $\phi^{-1} \circ T_0^x \circ \phi$ to approximate $T_1^x$. By this we mean that we should be able to predict as accurately as possible the result of some action $x$ in $O_1$ by translating our distribution for the initial state in $O_1$ to a distribution over $O_0$ (using the function $\phi$), predicting the result of action $x$ within $O_0$, and translating this result back to $O_1$ using $\phi^{-1}$. Similarly, we would like to use $O_1$ to predict the behavior of $O_0$.

Furthermore, we want to to optimize $\phi$ and $\phi^{-1}$ so that both models will make similar predictions about sensory data. Thus $A_0 \circ \phi$ should be close to $A_1$ and $A_1 \circ \phi^{-1}$ should be close to $A_0$.

To measure distance between two matrices, we treat the columns vectors as probability distributions and sum the Kullback-Leibler divergences of the columns. For two matrices $P$ and $Q$, let $D_KL(P||Q)$ be the sum of the Kullback-Leibler divergences of the columns. When calculating Kullback-Leibler divergence, we consider the columns of the $A$ and $T$ matrices to be the "true" distributions, while those depending on $\phi$ or $\phi^{-1}$ are regarded as the approximations.

So we choose $\phi$ and $\phi^{-1}$ to minimize the quantity

\begin{eqnarray*}
	\left(\sum_{x \in M} D_{KL} (T_1^x || \phi^{-1} T_0^x \phi)\right) + D_{KL}(A_1 || \phi^{-1} A_0 \phi) \\
	+ \left(\sum_{x \in M} D_{KL} (T_0^x || \phi T_1^x \phi^{-1})\right) + D_{KL}(A_0 || \phi A_1 \phi^{-1})
\end{eqnarray*}

Using a simple hill-climbing algorithm, we have tested our criterion on a simple example.

\section{Example: the long corridor}

The agent initially believes that it is standing in a corridor consisting of four discrete locations. The agent's actions are to move left or right. If the agent is already at the end of the corridor and attempts to move further in that direction, it will remain where it is. The agent can see whether it standing at the left end, the right end, or at neither end of the corridor. The agent's goal is to stand at the right end of the corridor.

\begin{center}
	\includegraphics[scale=1.0]{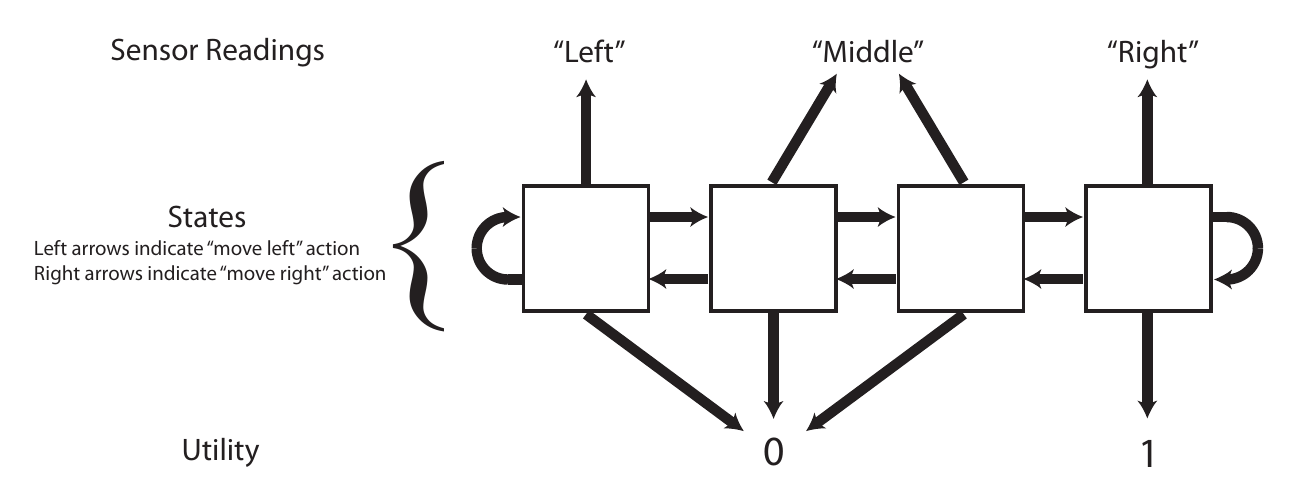}
\end{center}

Now the agent discovers that this ontology is incorrect; the corridor actually consists of five discrete locations. What, then, should the agent do? Intuitively, it seems most plausible that the agent should stand at the right end of the corridor. Stretching plausibility a bit, perhaps the agent should stand one step away from the right end of the corridor, since the corridor is longer than expected. Any other solution seems counterintuitive.

Our initial, four-state ontology $O_0$ can be represented in matrix form as follows:

\begin{equation}
T_0^L = \left( \begin{array}{cccc}
	1 & 1 & 0 & 0 \\
	0 & 0 & 1 & 0 \\
	0 & 0 & 0 & 1 \\
	0 & 0 & 0 & 0
\end{array} \right),
T_0^R = \left( \begin{array}{cccc}
	0 & 0 & 0 & 0 \\
	1 & 0 & 0 & 0 \\
	0 & 1 & 0 & 0 \\
	0 & 0 & 1 & 1
\end{array} \right)
A_0 = \left( \begin{array}{cccc}
	1 & 0 & 0 & 0 \\
	0 & 1 & 1 & 0 \\
	0 & 0 & 0 & 1
\end{array} \right)
\end{equation}

And the five-state ontology $O_1$ can be represented as:

\begin{equation}
T_1^L = \left( \begin{array}{ccccc}
	1 & 1 & 0 & 0 & 0 \\
	0 & 0 & 1 & 0 & 0 \\
	0 & 0 & 0 & 1 & 0 \\
	0 & 0 & 0 & 0 & 1 \\
	0 & 0 & 0 & 0 & 0
\end{array} \right),
T_1^R = \left( \begin{array}{ccccc}
	0 & 0 & 0 & 0 & 0 \\
	1 & 0 & 0 & 0 & 0 \\
	0 & 1 & 0 & 0 & 0 \\
	0 & 0 & 1 & 0 & 0 \\
	0 & 0 & 0 & 1 & 1
\end{array} \right)
A_1 = \left( \begin{array}{ccccc}
	1 & 0 & 0 & 0 & 0 \\
	0 & 1 & 1 & 1 & 0 \\
	0 & 0 & 0 & 0 & 1
\end{array} \right)
\end{equation}

By hill-climbing from random initial values, our program found several local optima. After 10 runs, our best result, to three significant figures, was:

\begin{equation}
\phi = \left( \begin{array}{ccccc}
	1 & 0 & 0 & 0 & 0 \\
	0 & 1 & 0.503 & 0 & 0 \\
	0 & 0 & 0.496 & 1 & 0 \\
	0 & 0 & 0 & 0 & 1 
\end{array} \right), 
\phi^{-1} = \left( \begin{array}{cccc}
	1 & 0.014 & 0.001 & 0 \\
	0 & 0.715 & 0 & 0\\
	0 & 0.270 & 0.283 & 0 \\
	0 & 0 & 0.715 & 0 \\
	0 & 0 & 0 & 1 
\end{array} \right)
\end{equation}

We can now make an interesting observation: $\phi \phi^{-1}$ is close to an identity matrix, as is $\phi^{-1} \phi$. Thus, after mapping from one ontology to the other, we can nearly recover our initial information.

\begin{equation}
\phi \phi^{-1}= \left( \begin{array}{cccc}
	1 & 0.014 & 0.137 & 0 \\
	0 & 0.851 & 0.142 & 0 \\
	0 & 0.134 & 0.856 & 0 \\
	0 & 0 & 0.001 & 1
\end{array} \right), 
\phi^{-1} \phi = \left( \begin{array}{ccccc}
	1 & 0.014 & 0.008 & 0.001 & 0 \\
	0 & 0.715 & 0.360 & 0 & 0 \\
	0 & 0.270 & 0.276 & 0.283 & 0 \\
	0 & 0 & 0.355 & 0.715 & 0 \\
	0 & 0 & 0.001 & 0  & 1
\end{array} \right)
\end{equation}

The matrix $\phi$ represents the following function mapping the 5-state ontology to the 4-state ontology:

\begin{center}
	\includegraphics[scale=0.667]{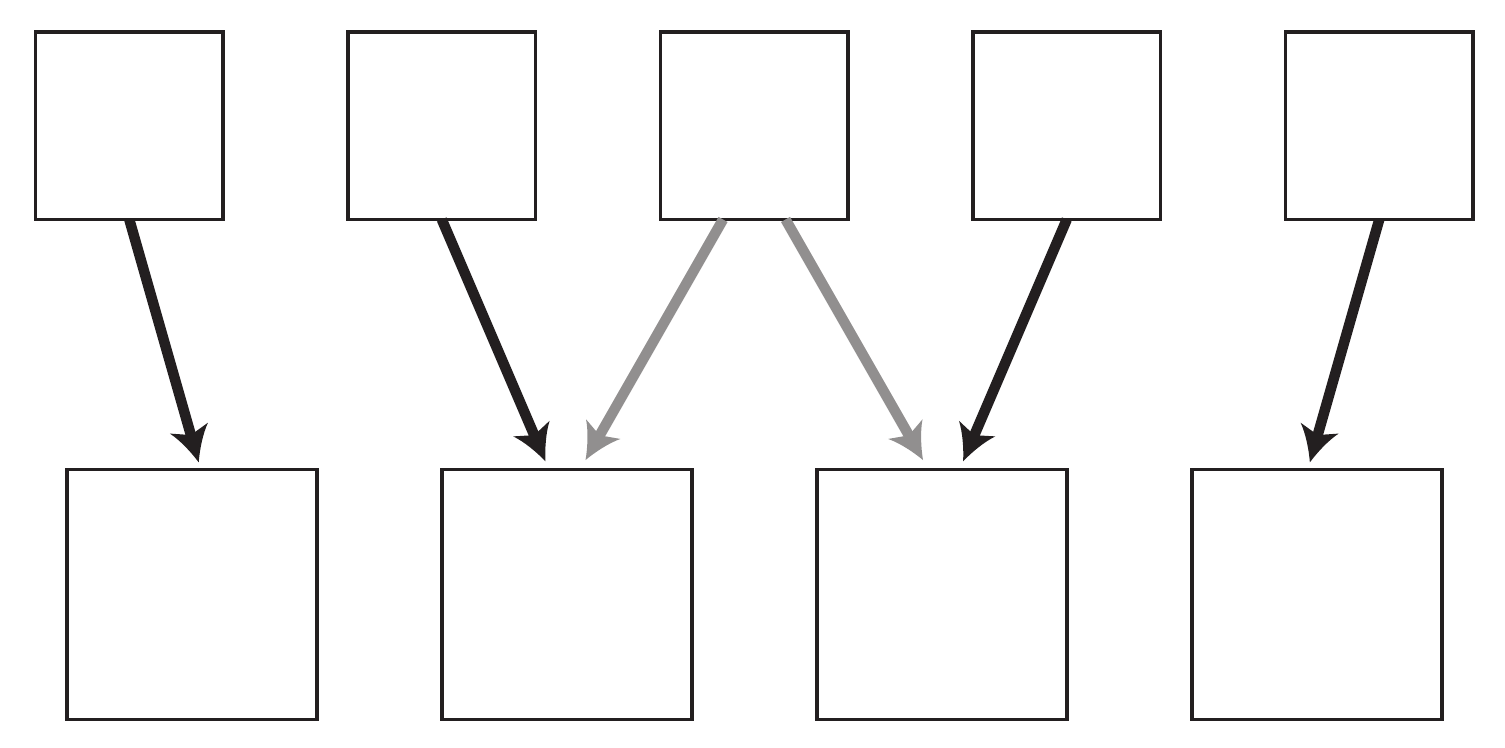}
\end{center}

The black arrows indicate near-certainty; the gray arrows indicate probabilities of about $\frac{1}{2}$.

If we compose $\phi$ with our utility function, we obtain a utility of 1 for the right square and a utility of 0 for the other squares, which agrees with our intuitions.

\section{Outlook}

Those wishing to extend our algorithm as presented may consider what to do when the agent's sensors or motors are replaced, how to deal with differently-sized time steps, how to deal with continuous models, and how to efficiently find mappings between larger, structured ontologies.

Furthermore, there remain difficult philosophical problems. We have made a distinction between the agent's uncertainty about which model is correct and the agent's uncertainty about which state the world is in within the model. We may wish to eliminate this distinction; we could specify a single model, but only give utilities for some states of the model. We would then like the agent to generalize this utility function to the entire state space of the model.

Human beings also confront ontological crises. We should find out what cognitive algorithms humans use to solve the same problems described in this paper. If we wish to build agents that maximize human values, this may be aided by knowing how humans re-interpret their values in new ontologies.

We hope that other thinkers will consider these questions carefully.

\section*{Acknowledgements}

Thanks to Roko Mijic, with whom I discussed these ideas in 2009, and to Steve Rayhawk, who gave extensive criticism on earlier versions of this paper.

\end{document}